\documentclass[10pt,twocolumn,letterpaper]{article}

\usepackage{wacv}

\usepackage{tikz}
\usepackage{listings}
\usepackage{tcolorbox}
\tcbuselibrary{listings,breakable}

\lstdefinestyle{prompt}{
    basicstyle=\ttfamily\tiny,
    breaklines=true,
    breakatwhitespace=false,
    columns=fullflexible,
    keepspaces=true,
    showstringspaces=false,
    frame=none,
    xleftmargin=0pt,
    xrightmargin=0pt,
}
\usepackage{pgfplots}
\usepackage{booktabs}
\usepackage{multirow}
\usepackage{subcaption}
\pgfplotsset{compat=1.18}
\usetikzlibrary{shapes.geometric, arrows.meta, positioning, calc, patterns}

\usepackage{xspace}

\definecolor{wacvblue}{rgb}{0.21,0.49,0.74}
\usepackage[pagebackref,breaklinks,colorlinks,allcolors=wacvblue]{hyperref}

\def\wacvPaperID{PRAW-9}
\def\confName{WACV}
\def\confYear{2026}

\title{Zero-Shot Product Attribute Labeling with Vision-Language Models:\\ A Three-Tier Evaluation Framework\thanks{© 2026 IEEE. Accepted to the 2026 IEEE/CVF Winter Conference on Applications of Computer Vision Workshops (WACVW). Personal use of this material is permitted. Permission from IEEE must be obtained for all other uses.}}

\author{Shubham Shukla\\
{\tt\small ss3469@cornell.edu}
\and
Kunal Sonalkar\\
{\tt\small kunal.sonalkar@nordstrom.com}
}

\begin{document}
\maketitle
\begin{abstract}
Fine-grained attribute prediction is essential for fashion retail applications 
including catalog enrichment, visual search, and recommendation systems. 
Vision-Language Models (VLMs) offer zero-shot prediction without task-specific 
training, yet their systematic evaluation on multi-attribute fashion tasks 
remains underexplored. A key challenge is that fashion attributes are often
\emph{conditional}. For example, ``outer\_fabric'' is undefined when no outer garment is
visible. This requires models to detect attribute applicability before attempting
classification. We introduce a three-tier evaluation framework that decomposes 
this challenge: (1) overall task performance across all classes (including NA class: suggesting attribute is not applicable) for all attributes, (2) attribute
applicability detection, and (3) fine-grained classification when attributes
are determinable. Using DeepFashion-MultiModal\footnote{\url{https://github.com/yumingj/DeepFashion-MultiModal}}, 
which explicitly defines NA (meaning attribute doesn't exist or is not visible) within attribute label spaces, 
we benchmark nine VLMs spanning flagship (GPT-5, Gemini 2.5 Pro), efficient 
(GPT-5 Mini, Gemini 2.5 Flash), and ultra-efficient tiers (GPT-5 Nano, 
Gemini 2.5 Flash-Lite) against classifiers trained on pretrained Fashion-CLIP embeddings on 5,000 images 
across 18 attributes. Our findings reveal that: (1) zero-shot VLMs achieve 
64.0\% macro-F1, a threefold improvement over logistic regression on pretrained Fashion-CLIP embeddings (21\% F1); (2) VLMs 
excel at fine-grained classification (Tier 3: 70.8\% F1) but struggle with 
applicability detection (Tier 2: 34.1\% NA-F1), identifying a key bottleneck; 
(3) efficient models achieve over 90\% of flagship performance at lower cost, offering practical deployment paths. This diagnostic framework 
enables practitioners to pinpoint whether errors stem from visibility detection 
or classification, guiding targeted improvements for production systems.
\end{abstract}

\section{Introduction}
\label{sec:intro}

Product attribute prediction is fundamental to modern retail systems. Accurate attributes enable catalog enrichment, visual search, personalized recommendations, and inventory management~\cite{ak2018efficient,guo2019imaterialist}. The fashion domain is particularly challenging, with fine-grained attributes spanning shape (neckline, waist accessories), fabric (cotton, denim, chiffon), and pattern (striped, floral, solid) that require both visual perception and domain knowledge.

Traditional approaches train specialized classifiers on labeled data~\cite{liu2016deepfashion,chia2022fashionclip}. While effective, these methods require significant annotation effort and struggle to generalize to new attributes. Vision-Language Models (VLMs) offer an alternative: zero-shot attribute prediction through natural language prompting, eliminating the need for task-specific training data.

Despite growing interest in VLMs for fashion retail applications, systematic evaluation methodologies remain underdeveloped. Standard evaluation using only macro-F1 provides a single aggregate score that cannot distinguish between different types of model failures. Specifically, it cannot reveal whether errors occur because the model cannot detect attribute applicability or because it struggles with fine-grained classification. In multi-attribute prediction tasks, many attributes are conditionally applicable. For instance, ``outer\_fabric'' is undefined when no outer garment (jacket, coat, cardigan) is visible, so models must first determine \textit{whether} an outer layer exists before classifying \textit{what} fabric it is made of. Current benchmarks conflate these distinct capabilities: (1) \textit{attribute applicability detection}, which tests whether the model can recognize when an attribute does not apply, and (2) \textit{fine-grained classification}, which tests whether the model can correctly classify visible attributes.

Inspired by comprehensive benchmarks like MMMU~\cite{yue2024mmmu} that rigorously evaluate VLMs across multiple disciplines, we observe that no analogous diagnostic framework exists for fashion attribute prediction that can pinpoint \textit{where} models fail rather than just \textit{how much}.

We address this gap with three contributions:

\begin{enumerate}[noitemsep,topsep=0pt]
    \item \textbf{Three-tier evaluation framework:} We decompose multi-attribute prediction into three diagnostic levels: (1) \textit{full task performance}, measuring overall accuracy across all classes including NA (not applicable); (2) \textit{attribute applicability detection}, measuring the model's ability to recognize when an attribute cannot be determined; and (3) \textit{classification given visibility}, measuring fine-grained discrimination when attributes are determinable. This decomposition pinpoints whether errors stem from visibility detection or classification.

    \item \textbf{Comprehensive benchmark:} We evaluate 9 state-of-the-art VLMs across model tiers, including flagship (GPT-5, Gemini 2.5 Pro), efficient (GPT-5 Mini, Gemini 2.5 Flash), and ultra-efficient (GPT-5 Nano, Gemini 2.5 Flash-Lite), against classifiers using pretrained Fashion-CLIP embeddings on 5,000 images across 18 attributes, with cost analysis.

    \item \textbf{Actionable deployment insights:} We find that zero-shot VLMs achieve 64\% F1 without any task-specific training, a threefold improvement over logistic regression trained on pretrained Fashion-CLIP embeddings (21\% F1), but struggle with attribute applicability detection (Tier 2: 34\% NA-F1). Efficient models achieve over 90\% of flagship performance at lower cost, enabling informed deployment decisions.
\end{enumerate}

Our framework and benchmark establish a foundation for comprehensive VLM evaluation on fine-grained attribute prediction in fashion retail, with potential extension to other e-commerce domains.

\section{Related Work}
\label{sec:related}

\paragraph{Fashion attribute prediction.}
The DeepFashion dataset~\cite{liu2016deepfashion} established benchmarks for clothing recognition with 1,000+ attribute labels. Subsequent work developed specialized architectures for multi-attribute prediction~\cite{ak2018efficient,chen2012describing}. Fashion-CLIP~\cite{chia2022fashionclip} fine-tuned CLIP on fashion data, achieving strong performance on retrieval and classification tasks.

\paragraph{LLMs and VLMs for fashion.}
Recent work has applied large language models to fashion attribute extraction. PAE~\cite{sinha2024pae} extracts product attributes from trend forecast PDFs using multimodal LLM processing, achieving 92.5\% F1-score on text extraction. However, their focus is on extracting attributes from marketing documents rather than evaluating visual understanding of product images against ground truth labels. Jiang \etal~\cite{jiang2024multilabel} use GPT-4 with few-shot learning to generate zero-shot classifiers for multi-label fashion classification, achieving 0.80 weighted F1-score on 18 aesthetic labels (e.g., ``feminine'', ``casual''). Their approach focuses on subjective style categories with few-shot training. In contrast, we perform purely zero-shot evaluation on objective physical attributes (fabric, pattern, shape) across 9 VLMs with a diagnostic three-tier framework and comprehensive cost analysis.

\paragraph{Vision-language models.}
Large VLMs including GPT-4V~\cite{openai2023gpt4v}, Gemini~\cite{team2023gemini}, and LLaVA~\cite{liu2023visual} have demonstrated impressive zero-shot visual understanding. Prior work evaluated VLMs on visual question answering~\cite{antol2015vqa}, image captioning~\cite{lin2014microsoft}, and visual reasoning~\cite{hudson2019gqa}. However, evaluation on structured multi-attribute prediction remains limited. Zhang \etal~\cite{zhang2024attribute} evaluated attribute comprehension in VLMs but focused on single attributes. We extend this to multi-attribute prediction with explicit handling of attribute applicability.

\paragraph{Evaluation methodologies and benchmarks.}
Decomposed evaluation has proven valuable in other domains. HELM~\cite{liang2022holistic} evaluates language models across multiple axes. Comprehensive benchmarks like MMMU~\cite{yue2024mmmu} have established rigorous evaluation standards for VLMs across 30 academic disciplines. For visual tasks, evaluation typically focuses on aggregate accuracy~\cite{goyal2017making}. We aim to bring similar rigor to fashion retail applications, creating a benchmark for fine-grained attribute prediction that can guide both research and deployment decisions. Our three-tier framework introduces applicability prediction as a distinct evaluation dimension, filling a gap in VLM evaluation methodology for tasks with conditional attribute visibility.

\paragraph{Foundation models for retail.}
Recent work has explored foundation models for fashion retail applications including product matching~\cite{bell2015learning}, virtual try-on~\cite{han2018viton}, and catalog enrichment~\cite{guo2019imaterialist}. The PRAW workshop series~\cite{praw2026} highlights growing interest in this intersection. Our work contributes a systematic evaluation framework and benchmark for zero-shot product attribute labeling, directly addressing the workshop's call for ``zero-shot data labeling with foundation models.''

\section{Dataset}
\label{sec:dataset}

We use the DeepFashion-MultiModal dataset~\cite{jiang2022text2human}, which provides fashion images with comprehensive attribute annotations. Our evaluation set contains 5,000 images with 18 attributes across three categories:

\begin{itemize}[noitemsep,topsep=0pt]
    \item \textbf{Shape attributes (12):} sleeve length, lower clothing length, neckline, socks, hat, glasses, neckwear, wrist wearing, ring, waist accessories, outer cardigan, upper covering navel
    \item \textbf{Fabric attributes (3):} upper fabric, lower fabric, outer fabric
    \item \textbf{Pattern attributes (3):} upper pattern, lower pattern, outer pattern
\end{itemize}

The data was stratified by product type during splitting. Train and development sets (9,000 samples: 7K train + 2K dev) were used to train baseline classifiers. All models (VLMs and baselines) were evaluated on the held-out test set of 5,000 images.

\paragraph{The NA (Not Applicable) Class.}
Of the 18 attributes, 16 include an NA class indicating the attribute is not visible, not applicable, or cannot be determined from the image. For example, ``outer\_fabric'' is undefined when no outer garment (jacket, coat, cardigan) is visible in the image. Two attributes, sleeve length and outer cardigan, always have determinable values and do not include an NA class. The presence of NA labels enables our three-tier evaluation framework, allowing us to separately assess attribute applicability detection and fine-grained classification. Table~\ref{tab:attributes} summarizes the attribute metadata.

\begin{figure}[t]
\centering
\begin{minipage}[c]{0.32\columnwidth}
\centering
\includegraphics[width=\textwidth]{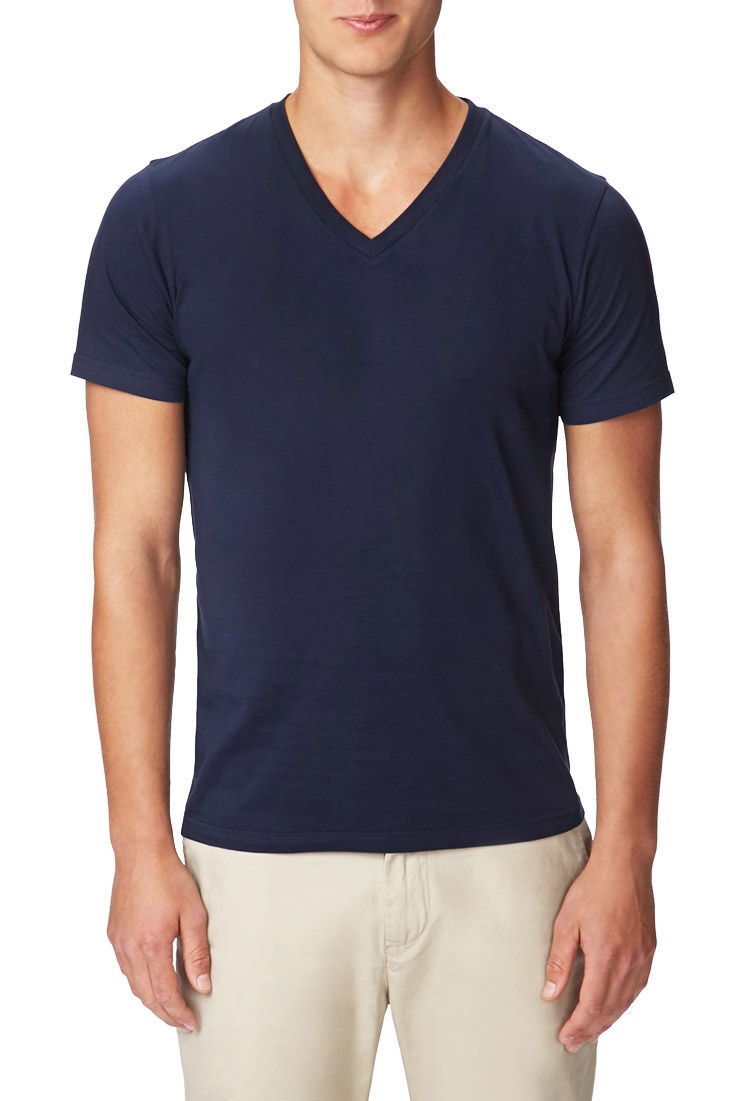}
\end{minipage}%
\hfill
\begin{minipage}[c]{0.65\columnwidth}
\footnotesize
\centering
\begin{tabular}{@{}ll@{}}
\toprule
\textbf{Attribute} & \textbf{Ground Truth} \\
\midrule
sleeve\_length & short-sleeve \\
lower\_clothing\_len & NA \\
socks, hat, glasses & NA \\
neckwear & no \\
wrist\_wearing, ring & no \\
waist\_accessories & NA \\
neckline & V-shape \\
outer\_cardigan & no \\
upper\_covering\_navel & yes \\
\midrule
upper\_fabric & cotton \\
lower\_fabric & cotton \\
outer\_fabric & NA \\
\midrule
upper\_pattern & pure color \\
lower\_pattern & pure color \\
outer\_pattern & NA \\
\bottomrule
\end{tabular}
\end{minipage}
\caption{Example test image with ground truth annotations. This outfit has 7 NA labels: lower clothing length not determinable from crop, socks/hat/glasses not visible, waist accessories unclear, and no outer garment.}
\label{fig:example}
\end{figure}

Figure~\ref{fig:example} shows a sample data point from our test set, illustrating how NA labels apply when attributes are not relevant to the garment type.

\begin{table}[t]
\centering
\small
\setlength{\tabcolsep}{3pt}
\begin{tabular}{@{}llcc@{}}
\toprule
\textbf{Attribute} & \textbf{Type} & \textbf{Classes} & \textbf{NA Label} \\
\midrule
sleeve\_length & shape & 4 & -- \\
lower\_clothing\_length & shape & 5 & \checkmark \\
neckline & shape & 7 & \checkmark \\
socks & shape & 4 & \checkmark \\
hat & shape & 3 & \checkmark \\
glasses & shape & 4 & \checkmark \\
neckwear & shape & 3 & \checkmark \\
wrist\_wearing & shape & 3 & \checkmark \\
ring & shape & 3 & \checkmark \\
waist\_accessories & shape & 4 & \checkmark \\
outer\_cardigan & shape & 2 & -- \\
upper\_covering\_navel & shape & 3 & \checkmark \\
\midrule
upper\_fabric & fabric & 8 & \checkmark \\
lower\_fabric & fabric & 8 & \checkmark \\
outer\_fabric & fabric & 8 & \checkmark \\
\midrule
upper\_pattern & pattern & 8 & \checkmark \\
lower\_pattern & pattern & 8 & \checkmark \\
outer\_pattern & pattern & 8 & \checkmark \\
\bottomrule
\end{tabular}
\caption{Attribute metadata. Classes include the NA class where applicable. Two attributes (sleeve\_length, outer\_cardigan) have no NA class as they are always determinable.}
\label{tab:attributes}
\end{table}

\section{Three-Tier Evaluation Framework}
\label{sec:method}

Standard evaluation metrics like macro-F1 aggregate model performance into a single score, obscuring important failure modes. In multi-attribute prediction tasks where attributes may not always be applicable (e.g., ``outer\_fabric'' is undefined when no outer garment is visible), models must solve two distinct sub-problems: (1) detecting whether an attribute is applicable, and (2) classifying the attribute value when applicable. We propose a three-tier evaluation framework that decomposes these capabilities for diagnostic analysis.

\subsection{Problem Formulation}

Given a product $x$ (consisting of an image and text description) and an attribute $a$ with label space $\mathcal{Y}_a = \{y_1, \ldots, y_k, \text{NA}\}$, a model predicts $\hat{y} \in \mathcal{Y}_a$. The task comprises 18 independent multiclass classification problems, one per attribute. The NA class indicates the attribute is not visible, not applicable, or cannot be determined from the product. Of our 18 attributes, 16 include an NA class while 2 (sleeve length, outer cardigan) always have a determinable value.

We compute metrics at three levels:
\begin{itemize}[noitemsep,topsep=0pt]
    \item \textbf{Class-level:} Per-class precision, recall, and F1 within each attribute
    \item \textbf{Attribute-level:} Macro-averaged F1 across classes: $F1_{attr} = \frac{1}{|C|} \sum_{c \in C} F1_c$
    \item \textbf{Model-level:} Mean of 18 attribute-level F1 scores: $F1_{model} = \frac{1}{18} \sum_{a=1}^{18} F1_{attr}^{(a)}$
\end{itemize}

\subsection{Tier Definitions}

\paragraph{Tier 1: Full Task Performance.}
Measures overall model performance on the complete classification task, treating NA as a valid class.

\begin{equation}
\text{Tier1-F1} = \frac{1}{|\mathcal{Y}_a|} \sum_{c \in \mathcal{Y}_a} F1_c
\end{equation}

where $F1_c$ is the F1 score for class $c$. This metric answers: \textit{How well does the model perform on the full attribute prediction task?}

\paragraph{Tier 2: Attribute Applicability Detection.}
Measures the model's ability to detect when an attribute is not applicable. We convert multiclass predictions to binary: NA $\rightarrow$ 1 (not applicable), any other class $\rightarrow$ 0 (applicable).

\begin{equation}
\text{Tier2-F1} = F1(\mathbf{1}[y = \text{NA}], \mathbf{1}[\hat{y} = \text{NA}])
\end{equation}

This tier captures two critical failure modes:
\begin{itemize}[noitemsep,topsep=0pt]
    \item \textbf{False visibility} (low NA-Recall): Model incorrectly predicts an attribute value when the attribute is actually not applicable
    \item \textbf{False NA} (low NA-Precision): Model incorrectly predicts NA when the attribute is clearly visible
\end{itemize}

\paragraph{Tier 3: Classification Given Applicability.}
Measures classification accuracy on samples where the attribute \textit{is} visible, testing fine-grained discrimination ability.

\begin{equation}
\text{Tier3-F1} = \frac{1}{|\mathcal{Y}_a|} \sum_{c \in \mathcal{Y}_a} F1_c \quad \text{s.t.} \quad y \neq \text{NA}
\end{equation}

Critically, NA remains in the prediction space. If the ground truth is ``cotton'' and the model predicts ``NA'', this is penalized as a false negative for cotton and a false positive for NA. This captures the failure mode where a model fails to classify a clearly visible attribute.

\subsection{Diagnostic Interpretation}

The three tiers enable systematic diagnosis of model behavior. Table~\ref{tab:diagnostic} summarizes the diagnostic patterns enabled by this framework.

\begin{table}[h]
\centering
\footnotesize
\begin{tabular}{@{}lp{4.2cm}@{}}
\toprule
\textbf{Pattern} & \textbf{Interpretation} \\
\midrule
Tier 1 $\approx$ Tier 3 & NA detection not a major factor \\
Tier 1 $\ll$ Tier 3 & Model struggles with NA detection \\
Low T2 NA-Recall & Predicts value when NA (false visibility) \\
Low T2 NA-Precision & Predicts NA when visible (false NA) \\
High T2, Low T3 & Knows \textit{when} to classify, not \textit{what} \\
Low T2, High T3 & Good discrimination, poor applicability \\
\bottomrule
\end{tabular}
\caption{Diagnostic patterns from three-tier analysis. These are misclassifications within the valid label space, distinct from hallucinations.}
\label{tab:diagnostic}
\end{table}

\subsection{Framework Visualization}

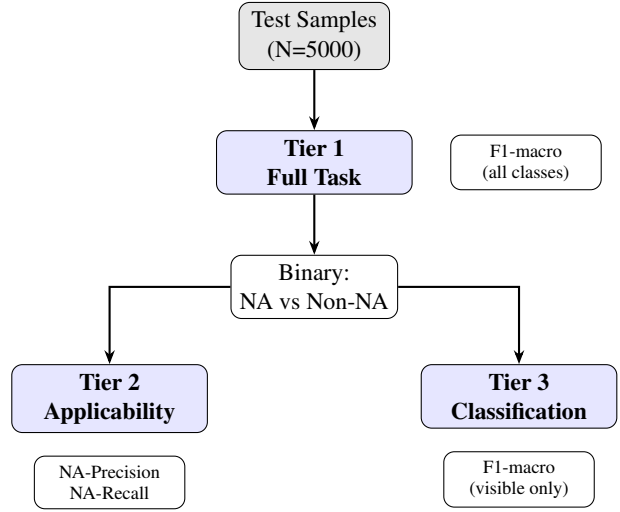
\begin{figure}[t]
\centering
\begin{tikzpicture}[
    node distance=0.8cm,
    box/.style={rectangle, draw, rounded corners, minimum width=2.0cm, minimum height=0.7cm, align=center, font=\small},
    tier/.style={rectangle, draw, fill=blue!10, rounded corners, minimum width=2.6cm, minimum height=0.8cm, align=center, font=\small\bfseries},
    arrow/.style={-{Stealth[scale=0.8]}, thick}
]

\node[box, fill=gray!20] (input) {Test Samples\\(N=5000)};

\node[tier, below=of input] (tier1) {Tier 1\\Full Task};
\node[box, right=0.5cm of tier1, font=\scriptsize] (t1m) {F1-macro\\(all classes)};

\node[box, below=of tier1] (split) {Binary:\\NA vs Non-NA};

\node[tier, below left=0.6cm and 0.3cm of split] (tier2) {Tier 2\\Applicability};
\node[box, below=0.3cm of tier2, font=\scriptsize] (t2m) {NA-Precision\\NA-Recall};

\node[tier, below right=0.6cm and 0.3cm of split] (tier3) {Tier 3\\Classification};
\node[box, below=0.3cm of tier3, font=\scriptsize] (t3m) {F1-macro\\(visible only)};

\draw[arrow] (input) -- (tier1);
\draw[arrow] (tier1) -- (split);
\draw[arrow] (split) -| (tier2);
\draw[arrow] (split) -| (tier3);

\end{tikzpicture}
\caption{Three-tier evaluation framework. Tier 1 evaluates all samples across all classes. Tier 2 measures binary NA detection. Tier 3 evaluates classification on samples where the attribute is visible (ground truth $\neq$ NA), with NA remaining in the prediction space.}
\label{fig:framework}
\end{figure}

Figure~\ref{fig:framework} illustrates the three-tier decomposition. All samples flow through Tier 1 evaluation. Tier 2 extracts binary applicability detection performance. Tier 3 filters to samples where the attribute is truly visible (ground truth $\neq$ NA), measuring classification accuracy on this subset while keeping NA in the prediction space.

\paragraph{Example.} For \texttt{upper\_fabric} with 8 classes (cotton, denim, leather, furry, knitted, chiffon, other, NA): Tier 1 computes F1-macro across all 8 classes on 5,000 samples. Tier 2 computes binary NA detection F1. Tier 3 computes F1-macro only on samples where upper garment is visible (e.g., 3,500 samples where ground truth $\neq$ NA), but a model predicting ``NA'' for a visible cotton shirt is penalized.

For attributes without an NA class (sleeve length, outer cardigan), Tier 1 and Tier 3 are identical, and Tier 2 does not apply.

\section{Experimental Setup}
\label{sec:experiments}

\subsection{Models}

We evaluate seven state-of-the-art VLMs organized by model tier, plus trained baseline classifiers and thinking ablations. Table~\ref{tab:models} summarizes the model groupings.

\begin{table}[t]
\centering
\footnotesize
\begin{tabular}{@{}llc@{}}
\toprule
\textbf{Group} & \textbf{Model} & \textbf{Vendor} \\
\midrule
\multirow{2}{*}{Flagship} & GPT-5 & OpenAI \\
& Gemini 2.5 Pro & Google \\
\midrule
\multirow{2}{*}{Efficient} & GPT-5 Mini & OpenAI \\
& Gemini 2.5 Flash & Google \\
\midrule
\multirow{2}{*}{Ultra-Efficient} & GPT-5 Nano & OpenAI \\
& Gemini 2.5 Flash-Lite & Google \\
\midrule
Non-Reasoning & GPT-4.1-mini & OpenAI \\
\midrule
Supervised & Fashion-CLIP + LR & Open Source \\
\midrule
\multirow{2}{*}{Thinking Ablation} & Gemini 2.5 Flash (Think) & Google \\
& Gemini 2.5 Flash-Lite (Think) & Google \\
\bottomrule
\end{tabular}
\caption{Model groupings by tier. VLMs are evaluated zero-shot; supervised baseline uses logistic regression on pretrained Fashion-CLIP embeddings. Thinking ablation uses 2048-token thinking budget.}
\label{tab:models}
\end{table}

\paragraph{Baseline: Fashion-CLIP + Logistic Regression.}
We use pretrained Fashion-CLIP~\cite{chia2022fashionclip} embeddings with logistic regression classifiers as our baseline. Fashion-CLIP is a CLIP model fine-tuned on fashion images. We train separate classifiers for each attribute using 5-fold cross-validation with GridSearchCV over regularization strength $C \in \{10^{-4}, \ldots, 10\}$. Training uses 9,000 samples (7K train + 2K dev) with class-balanced weighting. Note that our baseline uses frozen embeddings with a linear classifier, representing a minimal-adaptation approach. We intentionally use this design to isolate the value of pretrained representations: both the baseline and VLMs leverage frozen vision encoders without task-specific fine-tuning. Stronger supervised approaches, such as non-linear classifiers, data augmentation, or end-to-end fine-tuning, would likely narrow the gap but require additional labeled data and compute. Our baseline establishes a lower bound for supervised methods using pretrained embeddings; the threefold improvement demonstrates that zero-shot VLMs achieve strong absolute performance (64\% F1) without any task-specific training, a practical advantage when labeled data is scarce or annotation is costly.

We evaluate two baseline configurations:
\begin{itemize}[noitemsep,topsep=0pt]
    \item \textbf{Image-only:} 512-dimensional image embeddings
    \item \textbf{Multimodal:} 1024-dimensional concatenated image + text embeddings, where zero-shot Fashion-CLIP text embeddings are generated from product descriptions
\end{itemize}

\subsection{Multimodal Input}
\label{sec:multimodal}

Both VLMs and multimodal baselines receive the product image and a structured text description derived from filename metadata:

\begin{quote}
\small
\textit{``A [gender]'s [category] photographed from [view] view''}
\end{quote}

For example: \textit{``A women's dress photographed from front view''}. This provides contextual information about garment type and viewing angle.

\subsection{Evaluation Protocol}

\paragraph{Prompting.}
VLMs receive a zero-shot prompt defining the task: predict all 18 attributes with structured JSON output including the predicted class, confidence score (0--1), and brief reasoning. No in-context examples are provided. The complete prompt is shown in Figure~\ref{fig:prompt}.

\paragraph{Hallucination tracking.}
Predictions outside the valid label space (e.g., ``silk'' when valid classes are \{cotton, denim, ..., NA\}) are marked as hallucinations and assigned label $-1$. This allows us to separately analyze schema compliance.

\paragraph{Metrics.}
We report macro-F1, precision, and recall for each tier. All metrics use zero-division handling for edge cases. Model-level metrics are computed as the mean across all 18 attributes.

\paragraph{Safety filter handling.}
Gemini models blocked 14 images (0.28\%) due to content safety filters. These images were excluded from evaluation for all models to ensure comparable sample sets across model families.

\begin{figure*}[t!]
\centering
\begin{tcolorbox}[colback=gray!3, colframe=gray!60, title={\textbf{Zero-Shot System Prompt for VLM Evaluation}}, fonttitle=\small, boxrule=0.5pt, left=2pt, right=2pt, top=1pt, bottom=1pt]
\begin{lstlisting}[style=prompt]
You are an expert fashion attribute analyzer. Your task is to analyze fashion images and predict specific attributes about the clothing items shown.
## Task Description
Given an image of a fashion item and its text description, predict the values for 18 different fashion attributes organized into three categories: Shape, Fabric, and Pattern.
## Attribute Categories and Valid Values
### SHAPE ATTRIBUTES (12 total)
1. **sleeve_length** - Length of sleeves on upper clothing
   Valid values: "sleeveless", "short-sleeve", "medium-sleeve", "long-sleeve"
2. **lower_clothing_length** - Length of pants/skirts/shorts
   Valid values: "three-point", "medium short", "three-quarter", "long", "NA"
3. **socks** - Type of leg covering worn
   Valid values: "no", "socks", "leggings", "NA"
4. **hat** - Whether person is wearing a hat
   Valid values: "no", "yes", "NA"
5. **glasses** - Type of eyewear
   Valid values: "no", "sunglasses", "have a glasses in hand or clothes", "NA"
6. **neckwear** - Whether wearing necklace/scarf/tie
   Valid values: "no", "yes", "NA"
7. **wrist_wearing** - Whether wearing bracelet/watch
   Valid values: "no", "yes", "NA"
8. **ring** - Whether wearing a ring
   Valid values: "no", "yes", "NA"
9. **waist_accessories** - Accessories at waist
   Valid values: "no", "belt", "have a clothing", "NA"
10. **neckline** - Style of neckline on upper clothing
    Valid values: "V-shape", "square", "round", "standing", "lapel", "suspenders", "NA"
11. **outer_clothing_cardigan** - Whether outer layer is a cardigan
    Valid values: "yes", "no"
12. **upper_clothing_covering_navel** - Whether upper clothing covers navel
    Valid values: "no", "yes", "NA"
### FABRIC ATTRIBUTES (3 total)
13. **upper_fabric** - Fabric type of upper body clothing
    Valid values: "denim", "cotton", "leather", "furry", "knitted", "chiffon", "other", "NA"
14. **lower_fabric** - Fabric type of lower body clothing
    Valid values: "denim", "cotton", "leather", "furry", "knitted", "chiffon", "other", "NA"
15. **outer_fabric** - Fabric type of outer layer (jacket/coat)
    Valid values: "denim", "cotton", "leather", "furry", "knitted", "chiffon", "other", "NA"
### PATTERN ATTRIBUTES (3 total)
16. **upper_pattern** - Pattern on upper body clothing
    Valid values: "floral", "graphic", "striped", "pure color", "lattice", "other", "color block", "NA"
17. **lower_pattern** - Pattern on lower body clothing
    Valid values: "floral", "graphic", "striped", "pure color", "lattice", "other", "color block", "NA"
18. **outer_pattern** - Pattern on outer layer
    Valid values: "floral", "graphic", "striped", "pure color", "lattice", "other", "color block", "NA"
## Important Guidelines
1. **Use "NA"** when:
   - The item doesn't exist or is not visible in the image (e.g., no lower clothing visible, so lower_fabric = "NA")
   - The attribute doesn't apply to the item shown
   - The attribute cannot be determined from the image
2. **Be precise**: Choose the most specific value that matches what you see
3. **Provide reasoning**: For each attribute, explain in 1-2 sentences why you chose that specific value based on what you observe in the image
4. **Assign confidence scores**: Rate your certainty for each prediction on a scale of 0.0 to 1.0:
   Force yourself to use the full range. The confidence score reflects how certain you are about the ASSIGNED VALUE (including "NA").
   - **1.0**: Completely certain about the assigned value
     - For regular values: attribute is clearly visible and unambiguous
     - For "NA": completely certain the item doesn't exist or isn't applicable (e.g., dress clearly has no separate lower clothing)
   - **0.8-0.9**: Very confident about the assigned value
     - For regular values: attribute is clearly visible with minor ambiguity
     - For "NA": very confident the item doesn't exist, with only slight uncertainty
   - **0.6-0.7**: Moderately confident about the assigned value
     - For regular values: attribute is visible but has some uncertainty
     - For "NA": moderately confident it doesn't exist, but could be hidden/unclear
   - **0.4-0.5**: Uncertain about the assigned value
     - For regular values: difficult to determine, making an educated guess
     - For "NA": unclear if item exists or not (e.g., can't tell if there's a belt under clothing)
   - **0.2-0.3**: Very uncertain about the assigned value
     - For regular values: barely visible or highly ambiguous
     - For "NA": very unsure if item is absent or just not visible
   - **0.0-0.1**: Extremely uncertain - essentially guessing
5. **Output format**: For each attribute, provide three fields (value, reasoning, confidence). Return predictions as a JSON object with exactly this structure:
{"shape_attributes": {"sleeve_length": {"value": "<predicted_value>", "reasoning": "<explanation>", "confidence": <0-1>},
  "lower_clothing_length": {...}, ... (all 12 shape attributes)},
 "fabric_attributes": {"upper_fabric": {...}, ... (all 3 fabric attributes)},
 "pattern_attributes": {"upper_pattern": {...}, ... (all 3 pattern attributes)}}
Analyze the image carefully and provide your prediction in the exact JSON format specified above.
CRITICAL: Return ONLY the JSON object. No markdown code blocks, no preamble, no explanatory text before or after.
\end{lstlisting}
\end{tcolorbox}
\caption{Complete zero-shot prompt used for VLM evaluation. Models receive this prompt along with the product image and text description. No in-context examples are provided (purely zero-shot).}
\label{fig:prompt}
\end{figure*}

\section{Results}
\label{sec:results}

\subsection{Overall Performance Comparison}

Table~\ref{tab:main_results} presents the three-tier evaluation results for all models with bootstrap 95\% confidence intervals. The key finding is that \textbf{zero-shot VLMs substantially outperform logistic regression trained on pretrained Fashion-CLIP embeddings}: Gemini 2.5 Pro achieves 64.0\% macro-F1 [95\% CI: 58.1--69.4\%] compared to 20.8\% for the baseline, representing a threefold improvement without any task-specific training.

\begin{table}[t]
\centering
\resizebox{\columnwidth}{!}{%
\footnotesize
\begin{tabular}{@{}lccccc@{}}
\toprule
\textbf{Model} & \textbf{Tier 1 F1 (\%)} & \textbf{95\% CI} & \textbf{Tier 2 NA-F1 (\%)} & \textbf{Tier 3 F1 (\%)} & \textbf{Gap (\%)} \\
\midrule
\multicolumn{6}{l}{\textit{Supervised Baseline}} \\
Fashion-CLIP (Image) & 20.8 & -- & 17.1 & 17.2 & $-$3.6 \\
Fashion-CLIP (I+T) & 20.9 & -- & 17.7 & 17.2 & $-$3.7 \\
\midrule
\multicolumn{6}{l}{\textit{VLMs (Zero-shot)}} \\
Gemini 2.5 Pro & \textbf{64.0} & [58.1--69.4] & 34.1 & 65.4 & +1.4 \\
GPT-5 & 61.1 & [54.8--66.8] & \textbf{37.1} & 56.6 & $-$4.5 \\
Gemini 2.5 Flash & 59.9 & [53.6--65.7] & 22.0 & \textbf{70.8} & +10.9 \\
GPT-5 Mini & 58.0 & [52.0--64.0] & 22.7 & 65.7 & +7.7 \\
GPT-4.1-mini & 56.4 & [50.9--62.2] & 25.9 & 63.8 & +7.4 \\
GPT-5 Nano & 53.7 & [46.8--60.7] & 19.1 & 54.4 & +0.7 \\
Gemini 2.5 Flash-Lite & 53.2 & [45.7--60.6] & 23.4 & 58.5 & +5.3 \\
\midrule
\multicolumn{6}{l}{\textit{Thinking Ablation (2048 token budget)}} \\
Gemini 2.5 Flash (Think) & 59.8 & [53.4--65.7] & 20.9 & 68.1 & +8.3 \\
Gemini 2.5 Flash-Lite (Think) & 53.4 & [46.1--60.3] & 17.1 & 57.0 & +3.6 \\
\bottomrule
\end{tabular}%
}
\caption{Three-tier evaluation results with bootstrap 95\% confidence intervals (10,000 iterations). Gap = Tier 3 $-$ Tier 1 (positive indicates classification strength exceeds full-task performance). Best values in \textbf{bold}. Hallucination counts reported separately in Table~\ref{tab:hallucinations}.}
\label{tab:main_results}
\end{table}

\subsection{Performance by Model Tier}

Analyzing results by model tier (Table~\ref{tab:models}) reveals clear performance patterns:

\paragraph{Flagship models.} GPT-5 and Gemini 2.5 Pro achieve the highest Tier 1 F1 scores (61.1\% and 64.0\% respectively), with Gemini 2.5 Pro leading overall. However, GPT-5 shows superior NA detection (Tier 2: 37.1\% vs 34.1\%).

\paragraph{Efficient models.} GPT-5 Mini and Gemini 2.5 Flash provide 91--94\% of flagship Tier 1 performance at significantly lower cost. Gemini 2.5 Flash achieves the best Tier 3 classification (70.8\%) across all models, suggesting strong fine-grained discrimination despite weaker NA detection.

\paragraph{Ultra-efficient models.} Performance drops notably in this tier, with GPT-5 Nano (53.7\%) and Gemini 2.5 Flash-Lite (53.2\%). GPT-5 Nano shows the highest hallucination rate (2.78\%) across all models.

\paragraph{Non-reasoning model.} GPT-4.1-mini (56.4\%), an older non-reasoning model, outperforms the ultra-efficient tier while maintaining excellent schema compliance (only 2 hallucinations, 0.04\%). While this is a single data point, it suggests that explicit reasoning capabilities may not uniformly improve structured prediction performance across all model types.

\paragraph{Thinking ablation.} We evaluate whether extended reasoning improves performance by enabling a fixed 2048-token thinking budget on Gemini models. This allocates tokens for the model to ``think'' before generating structured outputs. We observe no statistically significant improvement in prediction accuracy: Gemini 2.5 Flash (Think) achieves 59.8\% vs.\ 59.9\% for standard Gemini 2.5 Flash ($-$0.1\%), while Gemini 2.5 Flash-Lite (Think) achieves 53.4\% vs.\ 53.2\% for standard Gemini 2.5 Flash-Lite (+0.2\%). However, thinking substantially improves schema compliance: Gemini 2.5 Flash-Lite (Think) produces only 3 hallucinations (0.06\%) compared to 35 (0.70\%) for standard Gemini 2.5 Flash-Lite, representing a 10$\times$ reduction. This suggests that explicit reasoning helps structured output generation but does not meaningfully improve attribute prediction accuracy.

\subsection{Tiered Analysis}

The three-tier framework reveals distinct failure modes across models:

\paragraph{Tier gap analysis.}
The gap between Tier 3 and Tier 1 indicates how much NA detection drags down overall performance. Gemini 2.5 Flash shows the largest positive gap (+10.9\%), meaning it excels at classification when attributes are visible but struggles significantly with NA detection (Tier 2: 22.0\%). Conversely, GPT-5 shows a negative gap (-4.5\%), indicating its NA detection (Tier 2: 37.1\%, the best) actually \textit{improves} its relative standing.

\paragraph{NA detection patterns.}
All VLMs struggle with applicability detection (mean Tier 2 NA-F1: 24.7\%). This suggests VLMs tend to ``over-detect'' by predicting specific attribute values even when attributes are not applicable, rather than conservatively predicting NA.

\subsection{Hallucination Analysis}

Table~\ref{tab:hallucinations} shows hallucination rates by model. The best-performing models (Gemini 2.5 Pro, GPT-4.1-mini) achieve near-perfect schema compliance with only 2 hallucinations each out of 5,000 images ($<$0.05\%). In contrast, GPT-5 Nano produces 139 hallucinations (2.8\%), suggesting smaller models struggle with structured output constraints.

\begin{table}[t]
\centering
\footnotesize
\setlength{\tabcolsep}{4pt}
\begin{tabular}{@{}lrrc@{}}
\toprule
\textbf{Model} & \textbf{Hall.} & \textbf{Rate (\%)} & \textbf{Cost (\$/5K)} \\
\midrule
Gemini 2.5 Pro & 2 & 0.04 & \$64.43 \\
GPT-4.1-mini & 2 & 0.04 & \$20.51 \\
Gemini 2.5 Flash-Lite (Think) & 3 & 0.06 & \$5.94 \\
GPT-5 Mini & 7 & 0.14 & \$20.76 \\
GPT-5 & 21 & 0.42 & \$131.22 \\
Gemini 2.5 Flash-Lite & 35 & 0.70 & \$2.91 \\
Gemini 2.5 Flash (Think) & 36 & 0.72 & \$28.38 \\
Gemini 2.5 Flash & 40 & 0.80 & \$14.86 \\
GPT-5 Nano & 139 & 2.78 & \$6.79 \\
\bottomrule
\end{tabular}
\caption{Hallucination counts and API costs. Hallucination rate = hallucinations / (18 attributes $\times$ available predictions).}
\label{tab:hallucinations}
\end{table}

\subsection{Cost-Performance Trade-offs}

\begin{figure}[t]
\centering
\begin{tikzpicture}
\begin{axis}[
    width=0.95\columnwidth,
    height=5.5cm,
    xlabel={Cost per 5K images (USD)},
    ylabel={Tier 1 F1 (\%)},
    xmode=log,
    log basis x=10,
    xmin=2, xmax=200,
    ymin=50, ymax=68,
    grid=both,
    grid style={line width=.1pt, draw=gray!20},
    major grid style={line width=.2pt,draw=gray!40},
    clip=false,
]
\node[circle, fill=blue, inner sep=1.5pt, label={[font=\tiny]above:{2.5 Pro}}] at (axis cs:64.43, 64.0) {};
\node[circle, fill=blue, inner sep=1.5pt, label={[font=\tiny]above:{2.5 Flash}}] at (axis cs:14.86, 59.9) {};
\node[circle, fill=blue, inner sep=1.5pt, label={[font=\tiny]below:{2.5 Flash-Lite}}] at (axis cs:2.91, 53.2) {};
\node[circle, draw=blue, inner sep=1.5pt, label={[font=\tiny]above:{Flash (Think)}}] at (axis cs:28.38, 59.8) {};
\node[circle, draw=blue, inner sep=1.5pt, label={[font=\tiny]above:{Lite (Think)}}] at (axis cs:5.94, 53.4) {};
\node[rectangle, fill=red, inner sep=1.5pt, label={[font=\tiny]above:{GPT-5}}] at (axis cs:131.22, 61.1) {};
\node[rectangle, fill=red, inner sep=1.5pt, label={[font=\tiny]above:{GPT-5 Mini}}] at (axis cs:20.76, 58.0) {};
\node[rectangle, fill=red, inner sep=1.5pt, label={[font=\tiny]below:{GPT-5 Nano}}] at (axis cs:6.79, 53.7) {};
\node[rectangle, fill=red, inner sep=1.5pt, label={[font=\tiny]below:{GPT-4.1-mini}}] at (axis cs:20.51, 56.4) {};
\end{axis}
\end{tikzpicture}
\caption{Cost-performance trade-off. \textcolor{blue}{$\bullet$} = Gemini, \textcolor{blue}{$\circ$} = Gemini (Think), \textcolor{red}{$\blacksquare$} = GPT. Gemini 2.5 Pro achieves best Tier 1 F1; Flash-Lite offers 22$\times$ cost reduction.}
\label{fig:cost_perf}
\end{figure}
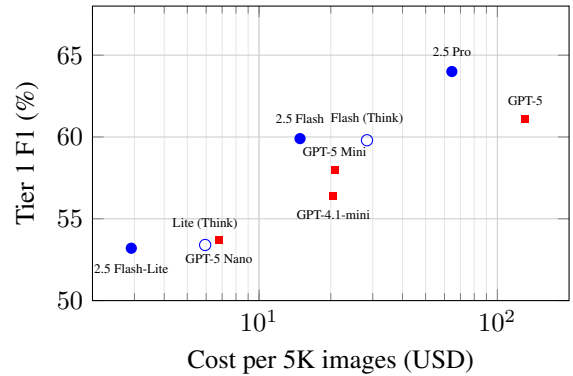

Figure~\ref{fig:cost_perf} and Table~\ref{tab:hallucinations} reveal significant cost variation. Gemini 2.5 Pro achieves the best Tier 1 F1 at \$64.43/5K images, while Gemini 2.5 Flash-Lite offers 83\% of that performance at just \$2.91/5K, providing a 22$\times$ cost reduction. GPT-5 is the most expensive at \$131.22/5K with lower performance than Gemini 2.5 Pro.

\subsection{Attribute-Level Performance}

Performance varies significantly across attribute types. Shape attributes achieve the highest mean F1 (e.g., outer\_cardigan: 78.2\% for Gemini 2.5 Pro), while fabric attributes are most challenging (upper\_fabric: 46.9\%). Pattern attributes fall in between. This hierarchy of shape $>$ pattern $>$ fabric holds across all models, suggesting fabric detection requires finer visual discrimination that current VLMs struggle with.

Table~\ref{tab:tier_breakdown} presents per-category breakdowns across all three evaluation tiers, averaging F1 scores within each attribute type.

\begin{table}[t]
\centering
\resizebox{\columnwidth}{!}{%
\footnotesize
\begin{tabular}{@{}ll|ccccccc@{}}
\toprule
& & \textbf{Gem} & \textbf{GPT} & \textbf{Gem} & \textbf{GPT} & \textbf{GPT} & \textbf{GPT} & \textbf{Gem} \\
\textbf{Tier} & \textbf{Type} & \textbf{Pro} & \textbf{5} & \textbf{Flash} & \textbf{Mini} & \textbf{4.1m} & \textbf{Nano} & \textbf{Lite} \\
\midrule
\multirow{3}{*}{T1} & Shape & \textbf{.70} & .65 & .63 & .61 & .60 & .60 & .57 \\
& Fabric & .49 & .51 & \textbf{.52} & .50 & .46 & .40 & .36 \\
& Pattern & .55 & .55 & \textbf{.56} & .54 & .51 & .42 & .54 \\
\midrule
\multirow{3}{*}{T2} & Shape & .24 & \textbf{.28} & .09 & .08 & .17 & .11 & .15 \\
& Fabric & .51 & \textbf{.51} & .43 & .47 & .42 & .33 & .38 \\
& Pattern & .51 & \textbf{.53} & .44 & .48 & .41 & .32 & .37 \\
\midrule
\multirow{3}{*}{T3} & Shape & .73 & .58 & \textbf{.79} & .72 & .71 & .56 & .64 \\
& Fabric & .45 & .46 & \textbf{.52} & .47 & .44 & .42 & .34 \\
& Pattern & .51 & .49 & \textbf{.56} & .53 & .47 & .44 & .53 \\
\bottomrule
\end{tabular}%
}
\caption{Per-category F1 breakdown across tiers. T1 = Full Task, T2 = NA Detection, T3 = Classification. Shape = 12 attrs, Fabric/Pattern = 3 each. Best per-row in \textbf{bold}.}
\label{tab:tier_breakdown}
\end{table}

\section{Discussion}
\label{sec:discussion}

\paragraph{When to use VLMs.}
Our results suggest VLMs are suitable for most fashion attribute prediction tasks, offering threefold improvement over logistic regression on pretrained embeddings while eliminating the need for labeled training data. VLMs are particularly effective when: (1) training data is limited or unavailable, (2) fine-grained classification is the primary objective, and (3) some tolerance for NA detection errors is acceptable. For applications requiring precise applicability detection (e.g., accessory presence), hybrid approaches combining VLM predictions with specialized NA classifiers may be beneficial.

\paragraph{Model selection guidance.}
Gemini 2.5 Pro offers the best overall performance (64.0\% F1) with excellent schema compliance (2 hallucinations). For cost-sensitive deployments, Gemini 2.5 Flash-Lite achieves 83\% of this performance at just 5\% of the cost. GPT-5 achieves the best NA detection (Tier 2: 37.1\%) but at 2$\times$ the cost of Gemini 2.5 Pro.

\paragraph{Deployment considerations.}
Two practical factors affect VLM deployment: (1) \textit{Safety filters}: Gemini models blocked 0.28\% of fashion images (14/5000) due to content filters. We excluded these images from evaluation across all models to maintain comparable test sets. (2) \textit{Hallucinations}: Even top models occasionally produce out-of-schema predictions; production systems need validation layers.

\paragraph{Limitations.}
Our evaluation has several limitations. First, our supervised baseline uses frozen embeddings with a linear classifier. This design choice enables fair comparison with zero-shot VLMs but likely underestimates the performance ceiling of supervised approaches; non-linear classifiers, data augmentation, or end-to-end fine-tuning could narrow the observed gap. Second, we benchmark only two proprietary VLM families (OpenAI, Google). Open-weights models such as LLaVA and Qwen-VL may exhibit different failure patterns, and our findings may not generalize across all VLM architectures. Third, we evaluate a single zero-shot prompt; chain-of-thought reasoning or few-shot examples may improve Tier~2 applicability detection, which we identify as the primary bottleneck. Fourth, results are limited to DeepFashion-MultiModal; evaluation on in-the-wild retail imagery would strengthen generalizability claims.

\section{Conclusion}
\label{sec:conclusion}

We introduced a three-tier evaluation framework for multi-attribute prediction that decomposes performance into full task metrics, applicability detection, and classification given visibility. Applying this framework to fashion attribute prediction, we benchmarked nine state-of-the-art VLMs and find that zero-shot VLMs outperform logistic regression trained on pretrained Fashion-CLIP embeddings by threefold (64\% vs.\ 21\% F1).

Our tiered analysis reveals a critical insight: VLMs excel at fine-grained classification when attributes are visible (Tier 3: 71\% F1) but struggle with attribute applicability detection (Tier 2: 34\% NA-F1). This diagnostic capability, which reveals \textit{where} models fail rather than just \textit{how much}, enables targeted improvements for deployment. Additionally, efficient models achieve over 90\% of flagship performance at lower cost, providing actionable deployment guidance.

Our framework and benchmark establish a foundation for comprehensive VLM evaluation on fine-grained attribute prediction in fashion retail, with potential extension to other e-commerce domains. Future work should explore stronger supervised baselines including non-linear classifiers and end-to-end fine-tuning to better characterize the performance ceiling, few-shot prompting and chain-of-thought reasoning to address the NA-detection bottleneck, open-weights VLMs (e.g., LLaVA, Qwen-VL) to assess generalizability beyond proprietary models, and evaluation on additional retail datasets with diverse imaging conditions.

{
    \small
    \bibliographystyle{ieeenat_fullname}
    \bibliography{main}

\begin{thebibliography}{21}
\providecommand{\natexlab}[1]{#1}
\providecommand{\url}[1]{\texttt{#1}}
\expandafter\ifx\csname urlstyle\endcsname\relax
  \providecommand{\doi}[1]{doi: #1}\else
  \providecommand{\doi}{doi: \begingroup \urlstyle{rm}\Url}\fi

\bibitem[Ak et~al.(2018)Ak, Lim, Tham, and Kassim]{ak2018efficient}
Kenan~E Ak, Joo~Hwee Lim, Jo~Yew Tham, and Ashraf~A Kassim.
\newblock Efficient multi-attribute similarity learning towards attribute-based
  fashion search.
\newblock In \emph{IEEE Winter Conf. Appl. Comput. Vis.}, pages 1671--1679,
  2018.

\bibitem[Antol et~al.(2015)Antol, Agrawal, Lu, Mitchell, Batra, Zitnick, and
  Parikh]{antol2015vqa}
Stanislaw Antol, Aishwarya Agrawal, Jiasen Lu, Margaret Mitchell, Dhruv Batra,
  C~Lawrence Zitnick, and Devi Parikh.
\newblock Vqa: Visual question answering.
\newblock In \emph{Int. Conf. Comput. Vis.}, pages 2425--2433, 2015.

\bibitem[Bell and Bala(2015)]{bell2015learning}
Sean Bell and Kavita Bala.
\newblock Learning visual similarity for product design with convolutional
  neural networks.
\newblock \emph{ACM Trans. Graph.}, 34\penalty0 (4):\penalty0 98:1--98:10,
  2015.

\bibitem[Chen et~al.(2012)Chen, Gallagher, and Girod]{chen2012describing}
Huizhong Chen, Andrew Gallagher, and Bernd Girod.
\newblock Describing clothing by semantic attributes.
\newblock In \emph{Eur. Conf. Comput. Vis.}, pages 609--623, 2012.

\bibitem[Chia et~al.(2022)Chia, Attanasio, Bianchi, Terragni, Magalh{\~a}es,
  Goncalves, Greco, and Tagliabue]{chia2022fashionclip}
Patrick~John Chia, Giuseppe Attanasio, Federico Bianchi, Silvia Terragni,
  Ana~Rita Magalh{\~a}es, Diogo Goncalves, Ciro Greco, and Jacopo Tagliabue.
\newblock Contrastive language and vision learning of general fashion concepts.
\newblock \emph{Scientific Reports}, 12:\penalty0 18958, 2022.

\bibitem[{Google DeepMind}(2023)]{team2023gemini}
{Google DeepMind}.
\newblock Gemini: A family of highly capable multimodal models.
\newblock \emph{arXiv preprint arXiv:2312.11805}, 2023.

\bibitem[Goyal et~al.(2017)Goyal, Khot, Summers-Stay, Batra, and
  Parikh]{goyal2017making}
Yash Goyal, Tejas Khot, Douglas Summers-Stay, Dhruv Batra, and Devi Parikh.
\newblock Making the v in vqa matter: Elevating the role of image understanding
  in visual question answering.
\newblock In \emph{IEEE Conf. Comput. Vis. Pattern Recog.}, pages 6904--6913,
  2017.

\bibitem[Guo et~al.(2019)Guo, Huang, Zhang, Srikhanta, Cui, Li, Scott, Adam,
  and Belongie]{guo2019imaterialist}
Sheng Guo, Weilin Huang, Xiao Zhang, Prasanna Srikhanta, Yin Cui, Yuan Li,
  Matthew~R Scott, Hartwig Adam, and Serge Belongie.
\newblock The imaterialist fashion attribute dataset.
\newblock In \emph{ICCV Workshops}, pages 3113--3116, 2019.

\bibitem[Han et~al.(2018)Han, Wu, Wu, Yu, and Davis]{han2018viton}
Xintong Han, Zuxuan Wu, Zhe Wu, Ruichi Yu, and Larry~S Davis.
\newblock Viton: An image-based virtual try-on network.
\newblock In \emph{IEEE Conf. Comput. Vis. Pattern Recog.}, pages 7543--7552,
  2018.

\bibitem[Hudson and Manning(2019)]{hudson2019gqa}
Drew~A Hudson and Christopher~D Manning.
\newblock Gqa: A new dataset for real-world visual reasoning and compositional
  question answering.
\newblock In \emph{IEEE Conf. Comput. Vis. Pattern Recog.}, pages 6700--6709,
  2019.

\bibitem[Jiang et~al.(2024)Jiang, Shah, Yeung, Zhu, Singh, and
  Goldenberg]{jiang2024multilabel}
Dongming Jiang, Abhishek Shah, Stanley Yeung, Jessica Zhu, Karan Singh, and
  George Goldenberg.
\newblock Multi-label classification for fashion data: Zero-shot classifiers
  via few-shot learning on large language models.
\newblock In \emph{International Conference on Knowledge Discovery and
  Information Retrieval (KDIR)}, pages 250--257, 2024.

\bibitem[Jiang et~al.(2022)Jiang, Yang, Qiu, Wu, Loy, and
  Liu]{jiang2022text2human}
Yuming Jiang, Shuai Yang, Haonan Qiu, Wayne Wu, Chen~Change Loy, and Ziwei Liu.
\newblock Text2human: Text-driven controllable human image generation.
\newblock \emph{ACM Trans. Graph.}, 41\penalty0 (4):\penalty0 1--11, 2022.

\bibitem[Liang et~al.(2022)Liang, Bommasani, Lee, Tsipras, Soylu, Yasunaga,
  Zhang, Narayanan, Wu, Kumar, et~al.]{liang2022holistic}
Percy Liang, Rishi Bommasani, Tony Lee, Dimitris Tsipras, Dilara Soylu,
  Michihiro Yasunaga, Yian Zhang, Deepak Narayanan, Yuhuai Wu, Ananya Kumar,
  et~al.
\newblock Holistic evaluation of language models.
\newblock \emph{arXiv preprint arXiv:2211.09110}, 2022.

\bibitem[Lin et~al.(2014)Lin, Maire, Belongie, Hays, Perona, Ramanan,
  Doll{\'a}r, and Zitnick]{lin2014microsoft}
Tsung-Yi Lin, Michael Maire, Serge Belongie, James Hays, Pietro Perona, Deva
  Ramanan, Piotr Doll{\'a}r, and C~Lawrence Zitnick.
\newblock Microsoft coco: Common objects in context.
\newblock In \emph{Eur. Conf. Comput. Vis.}, pages 740--755, 2014.

\bibitem[Liu et~al.(2023)Liu, Li, Wu, and Lee]{liu2023visual}
Haotian Liu, Chunyuan Li, Qingyang Wu, and Yong~Jae Lee.
\newblock Visual instruction tuning.
\newblock In \emph{Adv. Neural Inform. Process. Syst.}, 2023.

\bibitem[Liu et~al.(2016)Liu, Luo, Qiu, Wang, and Tang]{liu2016deepfashion}
Ziwei Liu, Ping Luo, Shi Qiu, Xiaogang Wang, and Xiaoou Tang.
\newblock Deepfashion: Powering robust clothes recognition and retrieval with
  rich annotations.
\newblock In \emph{IEEE Conf. Comput. Vis. Pattern Recog.}, pages 1096--1104,
  2016.

\bibitem[{OpenAI}(2023)]{openai2023gpt4v}
{OpenAI}.
\newblock Gpt-4v(ision) system card.
\newblock \url{https://openai.com/index/gpt-4v-system-card/}, 2023.

\bibitem[{PRAW Organizers}(2026)]{praw2026}
{PRAW Organizers}.
\newblock Physical retail ai workshop at wacv.
\newblock \url{https://grocery-vision.github.io/index.html}, 2026.

\bibitem[Sinha and Gujral(2024)]{sinha2024pae}
Apurva Sinha and Ekta Gujral.
\newblock Pae: Llm-based product attribute extraction for e-commerce fashion
  trends.
\newblock \emph{arXiv preprint arXiv:2405.17533}, 2024.

\bibitem[Yue et~al.(2024)Yue, Ni, Zhang, Zheng, Liu, Zhang, Stevens, Jiang,
  Ren, Sun, et~al.]{yue2024mmmu}
Xiang Yue, Yuansheng Ni, Kai Zhang, Tianyu Zheng, Ruoqi Liu, Ge Zhang, Samuel
  Stevens, Dongfu Jiang, Weiming Ren, Yuxuan Sun, et~al.
\newblock Mmmu: A massive multi-discipline multimodal understanding and
  reasoning benchmark for expert agi.
\newblock In \emph{IEEE Conf. Comput. Vis. Pattern Recog.}, pages 9556--9567,
  2024.

\bibitem[Zhang et~al.(2024)Zhang, Yang, Liu, Wang, He, Liang, and
  Ma]{zhang2024attribute}
Haiwen Zhang, Zixi Yang, Yuanzhi Liu, Xinran Wang, Zheqi He, Kongming Liang,
  and Zhanyu Ma.
\newblock Evaluating attribute comprehension in large vision-language models.
\newblock \emph{arXiv preprint arXiv:2408.13898}, 2024.

\end{thebibliography}
}

\end{document}